\documentclass{article}
\usepackage{amssymb}

\usepackage[final]{corl_2025} 

\usepackage{microtype}
\usepackage{graphicx}
\usepackage{subfigure}
\usepackage{booktabs}
\usepackage{makecell}
\usepackage{enumitem}

\definecolor{backcolour}{rgb}{0.97,0.97,0.97}

\usepackage{listings}
\lstdefinestyle{markdownstyle}{
    basicstyle=\ttfamily,
    backgroundcolor=\color{backcolour},   
    xleftmargin=0.05\textwidth,
    xrightmargin=0.05\textwidth,
    breakindent=0\dimen0,
    columns=flexible,
    showspaces=false,
    showstringspaces=false,
    breaklines=true,
    breakatwhitespace=true,
    breakautoindent=true,
}

\newcommand{\model}{\textsc{Articulate AnyMesh}}

\title{Articulate AnyMesh: Open-Vocabulary 3D Articulated Objects Modeling}

\author{
  Xiaowen Qiu*
  \And
  Jincheng Yang*
  \AND
  Yian Wang
  \And
  Zhehuan Chen
  \And
  Yufei Wang
  \And
  Tsun-Hsuan Wang
  \And
  Zhou Xian
  \And
  Chuang Gan
}

\begin{document}
\maketitle


\begin{abstract}
3D \textit{articulated} objects modeling has long been a challenging problem, since it requires to capture both accurate surface geometries and semantically meaningful and spatially precise structures, parts, and joints. Existing methods heavily depend on training data from a limited set of handcrafted articulated object categories (\textit{e.g.}, cabinets and drawers), which restricts their ability to model a wide range of articulated objects in an open-vocabulary context.
To address these limitations, we propose \model, an automated framework that is able to convert any rigid 3D mesh into its articulated counterpart in an open-vocabulary manner. Given a 3D mesh, our framework utilizes advanced Vision-Language Models and visual prompting techniques to extract semantic information, allowing for both the segmentation of object parts and the construction of functional joints.
Our experiments show that \model~can generate large-scale, high-quality 3D articulated objects, including tools, toys, mechanical devices, and vehicles, significantly expanding the coverage of existing 3D articulated object datasets. Additionally, we show that these generated assets can facilitate the acquisition of new articulated object manipulation skills in simulation, which can then be transferred to a real robotic system. Our Github website is \url{https://articulateanymesh.github.io/}.
\end{abstract}

\section{Introduction}
\label{submission}

\begin{figure*}[ht]
    \centering
    \includegraphics[width=0.85\linewidth, trim=0 160 0 120, clip]{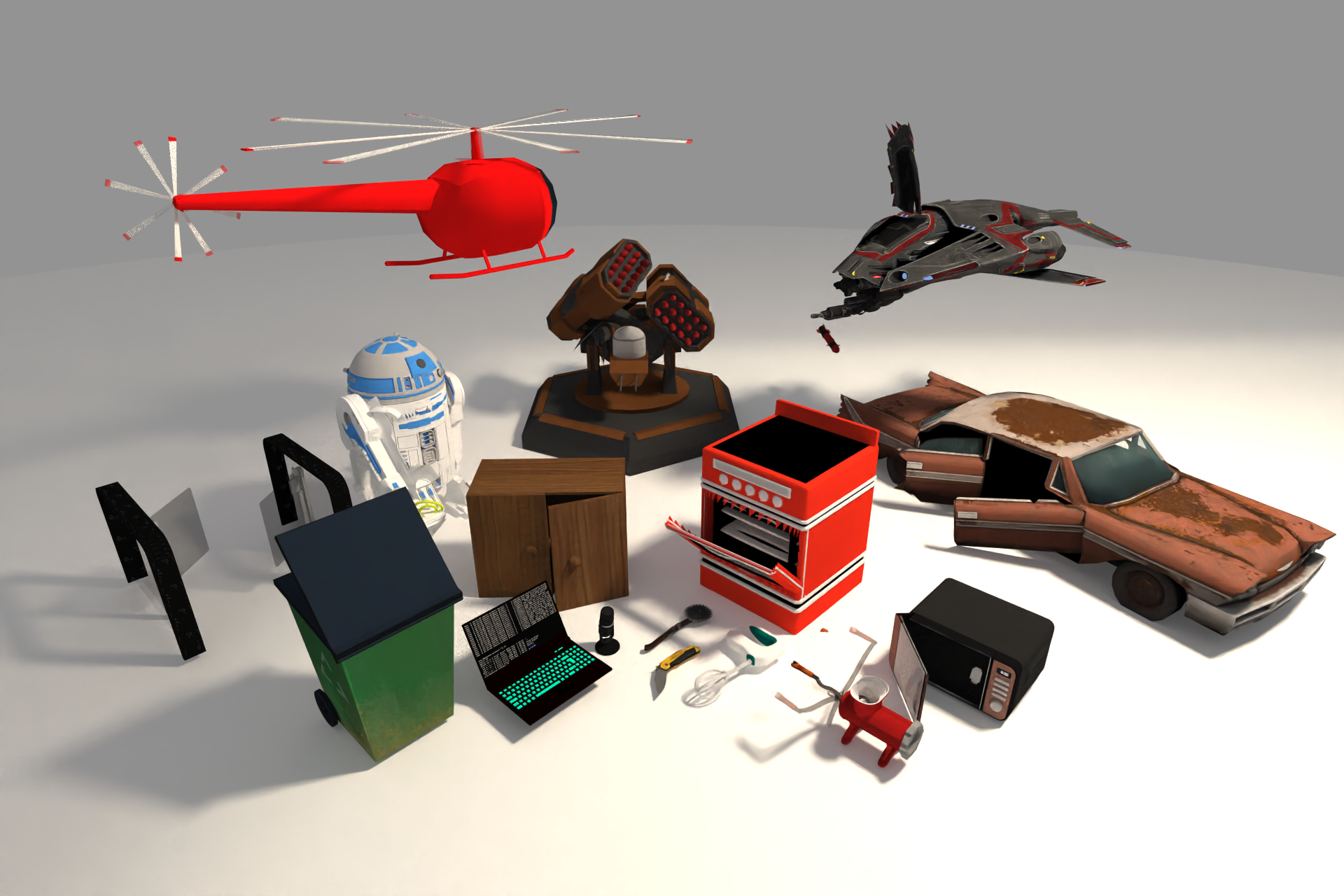}
    \vspace{-3mm}
    \caption{\textbf{\model} turns 3D meshes (\textit{e.g.} retrieved from Objaverse \cite{deitke2024objaverse}) into articulated objects. Our pipeline is capable of processing a wide range of objects, including daily necessities, furniture, vehicles and even fictional objects.}
    \label{fig:teaser}
\end{figure*}

The gathering of data on a large scale is an emerging research trend in embodied AI and robotics. Large foundation models built for robotics and embodied AI are extremely data-hungry, intensifying the need for large-scale data collection. Compared to collecting data in the real world, gathering data in simulations is significantly faster and more convenient, making it easier to capture various types of ground truth information such as physical states, segmentation masks, depth maps, etc. Existing works have explored many different aspects of how to carry out large-scale data collection in simulations, ranging from asset generation \citep{chen2024urdformer}, scene generation \citep{yang2024physcene,yang2024holodeck, wang2024architect}, task design \citep{wang2023robogen}, demonstration collection \citep{dalal2023imitating,ha2023scaling}, reward design \citep{ma2023eureka}, etc.

Despite these efforts, a key challenge remains: collecting diverse and realistic articulated objects essential for everyday life, which is vital for producing diverse data that can be generalized to real-world applications. One intuitive approach to achieve this is to use generative models.
While there has been substantial progress in 3D asset generation, few available methods are capable of addressing the demand for the collection of articulated objects. Most 3D generation approaches, utilizing a forward pass of a trained network \citep{long2024wonder3d, hong2023lrm, xu2024instantmesh, shi2023zero123++, shi2023mvdream}, or SDS loss optimization \citep{poole2022dreamfusion,wang2024prolificdreamer,qiu2024richdreamer}, produce only the surface of the object. These objects can only be manipulated as a whole body. For instance, a closet produced through these 3D generation methods cannot be opened or used to store clothes. Part-aware 3D generation approaches \citep{gao2019sdm,yang2022dsg,mo2019structurenet,wu2020pq,nakayama2023difffacto,koo2023salad,liu2024part123} generate 3D objects together with their part-specific details. Although these methods are more sensitive to structure, the objects produced are still restricted to whole-object manipulation for the lack of motion parameters. Articulated object creation approaches \citep{jiang2022ditto, liu2023paris, chen2024urdformer, lei2023nap, liu2024cage, mandi2024real2code, nie2023structure, gadre2021act} are capable of producing articulated objects with several interactive parts, demonstrating functionality. However, such methods require dense observation of a to be reconstructed articulated object in multiple joint states \citep{jiang2022ditto, liu2023paris, mandi2024real2code, huang2021multibodysync}, or are restricted to the limited-scale data and object categories used to train their network \citep{chen2024urdformer, lei2023nap, liu2024cage}. As a result, current methods struggle to automatically generate a wide variety of articulated objects, particularly those from underrepresented or absent categories in existing datasets \citep{xiang2020sapien, wang2019shape2motion, geng2023gapartnet}. 

Consequently, the collection of diverse articulated objects presents extra challenges compared to non-articulated 3D assets: (1) accurate semantic structures must accompany geometry and appearance, (2) articulation parameters such as joint orientation and position are required, and (3) the relatively small-scale articulated object datasets, compared to 3D object datasets, are insufficient to support the training of a generalizable model.
To overcome these challenges, we introduce \model, an automated pipeline that converts any 3D mesh into a corresponding articulated asset. In order to go beyond existing articulated object datasets, our pipeline leverages prior knowledge from visual and language foundation models \citep{achiam2023gpt, kirillov2023segment, rombach2022high} and generalizable geometric clues, rather than relying on existing labeled articulated object data. Our pipeline starts with a 3D mesh, either generated or handcrafted, followed by the stages of \textbf{Movable Part Segmentation}, \textbf{Articulation Estimation} and \textbf{Post-Processing}. 

In \textbf{Movable Part Segmentation}, the goal is to identify all movable parts and determine their semantics for subsequent articulation estimation. Recent open-vocabulary 3D part segmentation methods \citep{liu2023partslip, zhou2023partslip++, umam2024partdistill, xue2023zerops, yang2024sampart3d, liu2024part123} utilize an open-vocabulary segmentation model (namely SAM) and a multi-modal foundation model on rendered 2D images, integrating the 2D information to achieve 3D part segmentation. In this work, we adopt the pipeline proposed by PartSlip++ \cite{zhou2023partslip++} for 3D part segmentation.

After obtaining the 3D segmentation of non-fixed parts and their semantics, the next stage in our proposed pipeline focuses \textbf{Articulation Estimation}. Unlike previous approaches \citep{li2020category, huang2021multibodysync, zeng2024mars}, which rely on training datasets to predict articulation parameters. 
To overcome the limitations of existing datasets and extent to open-vocabulary manner, we extract geometric clues inherent in articulated objects instead of relying on learned models.  Specifically, the joint connecting two segmented parts is inherently related to the geometry of the connected area. For example, in cases where the connection follows a straight line, such as a laptop hinge or an open door, this line effectively represents the joint since the hinge structures align with it. We provide GPT-4o with such information through visual prompting, allowing it to leverage common-sense knowledge geometric reasoning to infer joint parameters. This approach allows generalization to unseen objects with accurate segmentation, and the performance will continue to improve as the vision-language model becomes more advanced.

At this stage, a functional articulated object has been created. However, if the input mesh is a generated or scanned surface mesh that contains only surface geometry and texture, the segmented 3D parts often suffer from occlusion and may exhibit holes. For example, in the closed state, only the outer surface of a microwave may be segmented, with the geometry of its internal cavity entirely missing. To address these issues and complete the pipeline, we incorporate an optional final stage: \textbf{Post-Processing}. Here, we leverage 3D generative models \citep{yang2025holopart, Meshy} to perform shape completion (filling holes and recovering missing geometry) and texture generation, making the pipeline self-contained for surface meshes.

We conducted experiments to compare joint parameter estimation accuracy with various baselines. Experimental results demonstrate that while our method matches state-of-the-art articulated object modeling methods for selected categories that they are trained on within PartNet-Mobility dataset \cite{xiang2020sapien}, it also exhibits the capacity to model a broader range of articulated objects beyond these categories. 
We summarize our contributions as follows: 
\begin{itemize}[itemsep=0pt, topsep=2pt]
    \item We introduce \model, a pipeline capable of creating diverse, realistic and complex articulated objects in an open-vocabulary manner.
    \item We propose a novel articulation parameter estimation method that leverages geometric clues and a tailored visual prompting method to estimate joint parameters in an open-vocabulary manner. 
    \item We conducted extensive experiments to evaluate the performance of our pipeline. We compared our method with others in joint parameter estimation and outperformed previous approaches in both in-domain and out-of-domain settings. We also carried out manipulation policy learning experiments to demonstrate the usefulness of the data collected by our pipeline.

\end{itemize}

\section{Related Work}
\subsection{Articulated object modeling} 
One line of work focuses on the perception end, including reconstruction of articulated objects from observations, joint parameter estimation, etc. \cite{jiang2022ditto} and \cite{huang2021multibodysync} train a network to reconstruct an articulated object from input multi-frame point clouds. Some other works \citep{hsu2023ditto, gadre2021act, nie2023structure, wang2022adaafford} learn joint parameters from active interaction. \cite{liu2023paris}, \cite{weng2024neural}, \cite{wei2022self} and \cite{mandi2024real2code} reconstruct the geometry and joint parameters of an articulated object from multiview images obtained at distinct states of the articulated object. \cite{chen2024urdformer} reconstructs articulated objects from a single real-world RGB image through a network trained on large-scale synthetic data. \citep{li2020category, jiang2022opd, sun2024opdmulti, yan2020rpm, liu2022toward, liu2023self, geng2023gapartnet} estimates joint parameters and part segments given the poincloud or image of an articulated object. However, these approaches either demand extensive observation of the same object or rely on existing articulated object data to train neural networks, with limited ability to generalize across only a small range of categories.

Alongside perception works, generation works including \cite{lei2023nap}, \cite{liu2024cage} and \cite{liu2024singapo}, represents articulated objects as graphs and employs a diffusion model to fit the distribution. The shape, position, and semantic information of the parts are encoded in the vertices, while joint parameters are stored in the edges.  Similar to perception methods, these generation methods also depend on existing articulated object datasets to train their diffusion models and are unable to generalize beyond the training data. 

Overall, none of the existing methods can automatically create articulated objects in an open-vocabulary manner. 
\nocite{liu2024survey}

\subsection{Part aware 3D generation} Instead of generating an 3D object as a whole, part aware 3D generation methods generate objects with part-level information. \cite{gao2019sdm} learns two separate VAEs to model global structural information and detailed part geometries. \cite{yang2022dsg} encodes 3D objects into a representation that disentangles structure and geometry. \cite{mo2019structurenet} utilizes a graph VAE to encode shapes into hierarchical graph representations. \cite{wu2020pq} applies a Seq2Seq generative network for part assembly and generation. \cite{nakayama2023difffacto} independently models part geometry and global part transformation, and conditioned on both of them, a diffusion model synthesizes the final shape. \cite{koo2023salad}, on the other hand, generates part geometry latents with one diffusion model and synthesizes the global geometry with another diffusion model conditioned on these latents. \cite{liu2024part123} learns a Neus model from generated multi-view images and corresponding 2D segmentation maps provided by \cite{kirillov2023segment}, and extracts 3D segmentation masks using a clustering algorithm. \cite{chen2024partgen} starts from multi-view images, applies diffusion-based networks for consistent part segmentation, completes each part across views, and reconstructs 3D parts using a pre-trained model.. 

While these methods can produce shapes with plausible structures and part geometries, they frequently depend on object datasets with part-level annotations and fail to generalize beyond the datasets used for training. Furthermore, they do not generate articulation parameters for individual parts, causing the generated parts to be individually non-manipulatable in simulation, which limits their applicability in downstream tasks for robot learning.

\section{Method}

\begin{figure*}[ht]
    \centering
    \includegraphics[width=\linewidth]{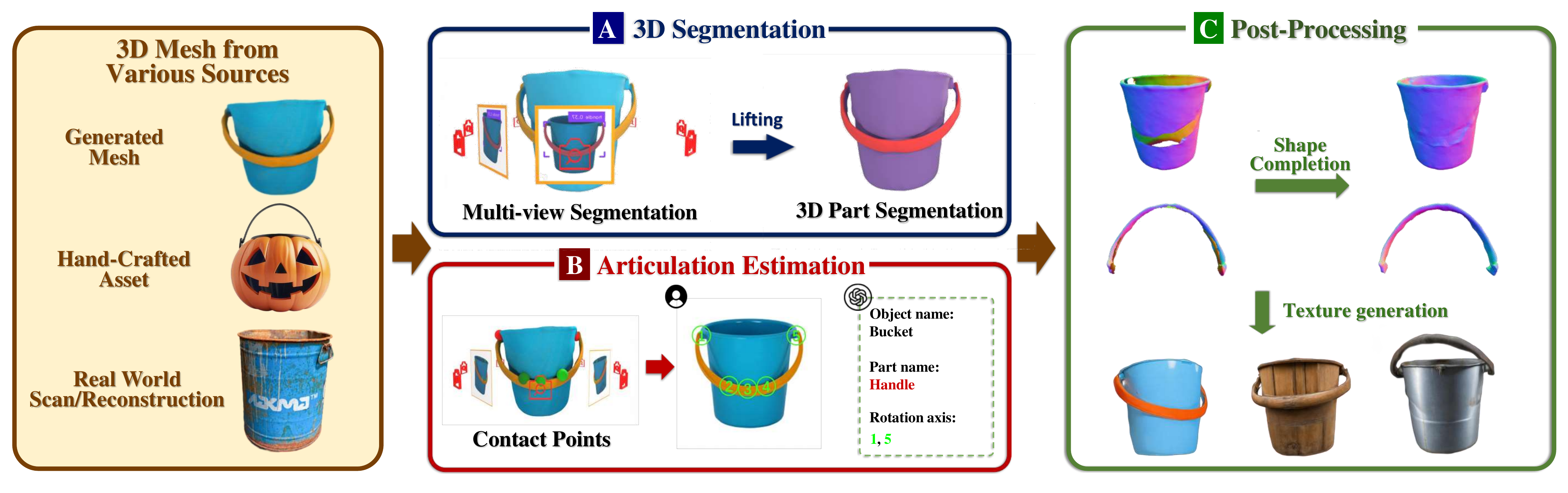}
    \vspace{-3mm}
    \caption{\model~converts 3D meshes from various sources to high quality articulated objects via three main parts: \textbf{A.} Movable Part Segmentation, \textbf{B.} Articulation Estimation, \textbf{C.} Post-Processing.}
    \label{fig:pipeline}
    \vspace{-3mm}
    
\end{figure*}

\subsection{Overview}
Our pipeline converts any mesh into its articulated counterpart, including hand-crafted meshes with part geometry, surface meshes generated through text-to-3D or image-to-3D methods and meshes reconstructed from real-world objects. The process involves three main steps, including \textbf{Movable Part Segmentation}, \textbf{Articulation Estimation} and \textbf{Post-Processing}, as depicted in Figure~\ref{fig:pipeline}. Specifically, the \textbf{Movable Part Segmentation} step employs the method proposed by PartSlip++ \cite{liu2023partslip} to segment movable parts from the input mesh.  \textbf{Articulation Estimation} extracts the articulation parameters for each movable part by leveraging geometric cues and prior knowledge from visual and language foundation models. The optional \textbf{Post-Processing} stage enhances geometry and texture using an off-the-shelf 3D shape completion model \cite{yang2025holopart} and an off-the-shelf texture generation model \cite{Meshy}.

\subsection{Movable Part Segmentation via VLM assistant}
As shown in Figure~\ref{fig:pipeline}A, this step segments all movable parts from a 3D mesh and identifies their semantics in an open-vocabulary manner. We primarily adopt PartSlip++ \cite{zhou2023partslip++} for segmentation. Given a 3D object and a list of part labels, PartSlip++ applies an open-vocabulary grounding model \citep{ren2024dino, li2022grounded} and SAM to 2D rendered images to extract bounding boxes and segmentation masks for the relevant semantic labels. These multi-view 2D detections are then fused to produce the final 3D part segmentation. In our pipeline, the label list is generated by feeding an image of the input mesh into a VLM to identify all movable parts.

\subsection{Articulation Estimation via Geometry-aware Visual Prompting}
Achieving an open-vocabulary scope for articulation estimation requires going beyond the limited-scale datasets of articulated objects.
To address this, we propose a \textit{Geometry-aware Visual Prompting} approach that generalizes across a diverse range of articulated objects by leveraging shared features in geometry, mechanical structure, and functionality. Specifically, we observe that joints connecting two parts are strongly relevant to the geometry in areas where the parts are in close proximity, which we will refer to as the \textit{connecting area}. 

Denote the point clouds of two neighboring parts $a$ and $b$ as $P_a$ and $P_b$ respectively. To define the connecting area, we select the points in $P_a$ whose distance to the closest point in $P_b$ is below a predefined threshold and vice versa for points in $P_b$. The selected points from $P_a$ and $P_b$ together form the connecting area between the parts. If no point is selected, the threshold is doubled, and the process is repeated until a connecting area is identified. We focus primarily on two dominant joint types: prismatic joints and revolute joints. These are estimated using distinct methods tailored to their respective mechanical structures.

\textbf{Revolute joints} 
connect two parts of an articulated object through physical hinges or other similar hinge-like mechanisms, such as the lid of a cardboard box. The rotation axis of a revolute joint aligns with the length of its hinge, meaning that identifying hinges in 3D space allows for straightforward estimation of the rotation axis. Since a hinge physically connects two parts, it is typically located within the region where the parts are joined. Driven by this observation, we propose the following visual prompting procedure. First, we divide the connecting area into multiple clusters, with the cluster centers serving as candidate points. 
These candidate points are then projected onto 2D rendered images of the input mesh and labeled with numbers, as illustrated in Figure~\ref{fig:pipeline}B.
This image is used to prompt GPT4o to select two or more points that define the hinge, thereby establishing the rotation axis. The rendering viewpoint is chosen to maximize the number of pixels of the part under consideration. K-means is run multiple times with varying cluster counts, and the final count is selected to maximize the number of candidate points while avoiding overlapping labels on the rendered image.

In some cases, a hinge may not be positioned on the object's surface, resulting in only one end of the rotation axis identifiable on the object's surface, such as with knobs. In these cases, we assume the rotation axis to be perpendicular to the plane fitted to the connected area. The axis is then determined by the normal vector of this plane and the single selected point. When the hinge mechanism is hidden beneath the surface, the connecting structure cannot be assumed to be entirely within the connecting area. Thus, in addition to candidate points from the connecting area, we sample additional points from the part surface. Specifically, we use the $l_0$-cut pursuit \cite{landrieu2017cut} algorithm to generate super points from the part's point cloud, using point normals and color as features, and derive candidate points from these super points. This ensures the new candidate points are distributed across different geometric and textured regions of the part.

\textbf{Prismatic joints} 
connect two parts of an articulated object through sliding mechanisms, enabling linear motion along a single axis. If the input mesh includes physically accurate inner geometry, sampling a collision-free sliding direction is likely to yield an accurate prismatic joint axis. However, most meshes, particularly surface meshes, lack the ideal geometry needed for this collision-based approach. Additionally, the sliding mechanism is often concealed beneath the object's surface, making it less identifiable compared to hinge structures for revolute joints and making the strategy used for revolute joints impractical for prismatic joints.

Based on their functionality, prismatic joints can be categorized into two types: 

1. \textbf{Inward \& Outward sliding joints}: These joints allow the child component to slide in and out of the object it is connected to, as seen in parts like drawers and push buttons. Although the translation direction may vary, the motion to pull the part out is predominantly outward from the object. Consequently, the sliding direction is characterized by the normal vector of the plane fitted to the connecting area.

2. \textbf{Surface sliding joints}:
These joints enable the child component to slide along the surface of the object, as seen in parts like sliding windows and toaster levers. For such joints, the potential translation direction is constrained to two dimensions along the surface. This allows us to annotate a 2D rendered image with arrows and prompt GPT-4 to select the most appropriate direction.

Given an articulated object and its segmented parts, we prompt GPT-4 to classify all prismatic joints into these two categories. Based on the classification, the translation directions of the prismatic joints are then estimated accordingly.

\subsection{Geometry and Texture Post-Processing}

After completing 3D segmentation and joint estimation, we obtain an articulated object with the correct articulation structure. However, when the input is a surface mesh, the result often remains incomplete. Parts segmented from surface meshes—lacking internal geometry—are prone to occlusions, resulting in incomplete meshes with holes. To address this, we introduce an optional post-processing step for surface meshes (primarily generated ones), where off-the-shelf 3D generation models are used for shape completion and texture generation, as illustrated in Section C of Figure~\ref{fig:pipeline}.

For shape completion, we use HoloPart\cite{yang2025holopart}, which takes a mesh with 3D segmentation labels as input and outputs mesh parts with completed geometry. As shown in the pipeline figure, HoloPart repairs issues such as holes in the bucket base (caused by occlusion from the handle) and a broken handle. For texture generation, we use Meshy\cite{Meshy} to apply textures to the completed parts. As shown in the figure, Meshy can generate high-quality textures with diverse materials, such as wood, plastic, and metal.

\section{Experiments} \label{experiments_section}
\textbf{Implementation}
\nocite{Genesis}
In the \textbf{movable part segmentation} stage, we use Partslip++\cite{zhou2023partslip++} for segmentation and replace the grounding model with a pipeline that uses set-of-mark visual prompting\cite{yang2023set} to enable the VLM to perform grounding for better performance. In addition, GPT4o is prompted to extract part labels of the input mesh and some of its joint configurations like joint type, parent link, joint limit, etc.
In \textbf{articulation estimation} stage, we visually prompt GPT4o for the rest of the joint configurations. 
In the \textbf{Post-Processing} stage, we employ HoloPart \cite{yang2025holopart} for 3D shape completion and Meshy \cite{Meshy} for texture generation.

\textbf{Baselines} 
To the best of our knowledge, this is the first work to tackle the challenge of converting meshes into articulated objects in an open-vocabulary manner. As a result, no existing baseline directly processes the same input and output as \model. Therefore, we compare against prior works that perform joint parameter estimation only on object categories seen during training.

URDFormer \cite{chen2024urdformer} takes a front-view image as input and sequentially outputs part segmentation, joint parameters, and part geometry. Real2Code \cite{mandi2024real2code} takes multi-view images as input and performs part segmentation, shape completion, and joint parameter estimation through code generation. For these two methods, we compare \model on the PartNet-Mobility dataset \cite{xiang2020sapien}. Additional comparisons with estimation methods such as ANCSH\cite{li2020category}, OPD\cite{jiang2022opd}, and OPD-Multi\cite{sun2024opdmulti}, as well as generative methods like NAP\cite{lei2023nap} and CAGE\cite{liu2024cage}, are provided in Appendix section \ref{sec_quanti}. We also provide an ablation study analyzing the impact of different vision-language models (VLMs), including GPT, Gemini, and Claude, in Appendix section \ref{sec_abla_vlm}.

\textbf{Metrics} 
For articulation parameter estimation, we evaluate joint directional accuracy using angular error and joint positional accuracy using the distance between lines.

\subsection{Quantitative Experiments}
We compare \model to URDFormer and Real2Code on the PartNet-Mobility dataset. To ensure a comprehensive evaluation, we assess joint estimation performance on both in-domain (ID) and out-of-domain (OOD) object categories. The in-domain categories include the five object types used to train URDFormer: oven, fridge, cabinet, washer, and dishwasher. For Real2Code, we fine-tune a CodeLlama model on objects from the same categories. We align the inputs of all methods as closely as possible to ensure a fair comparison: URDFormer takes a front-view image and is additionally provided with ground-truth part bounding boxes to predict joint parameters per part; Real2Code takes ground-truth part meshes and performs joint estimation via code generation; \model takes the ground-truth mesh, performs part segmentation, and estimates joint parameters via visual prompting. Results in table \ref{tab:former_real2code} show that although URDFormer and Real2Code perform just slightly worse on in-domain categories, their performance drops significantly on out-of-domain objects. In contrast, \model maintains low error across both ID and OOD settings. This highlights the key advantage of our approach: it is not constrained by the biases of existing articulated object datasets and generalizes well beyond them.

\begin{figure*}[ht]
    \centering
    \includegraphics[width=0.9\linewidth]{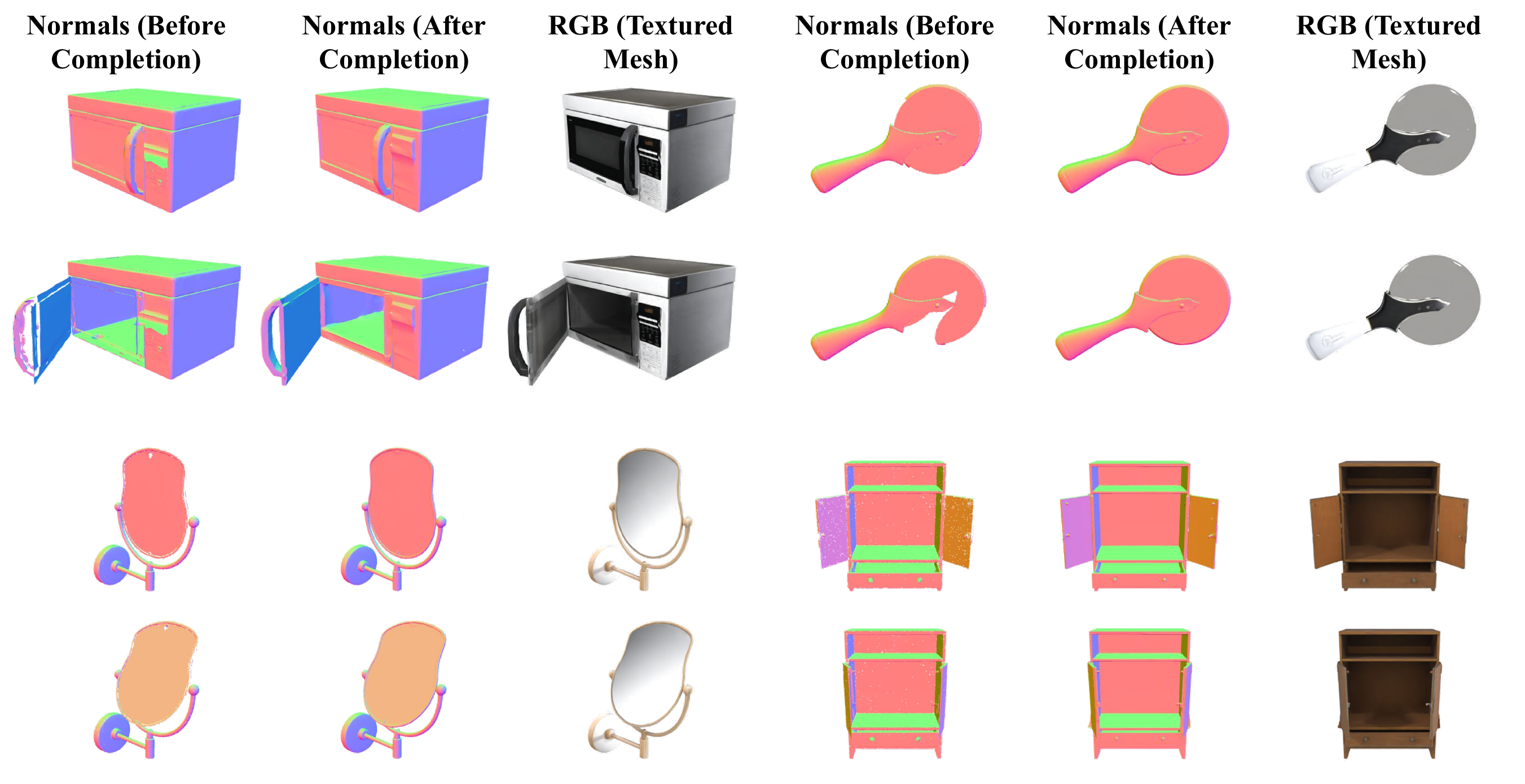}
    \caption{Results of articulated object generation. Each example shows (from left to right): the raw surface mesh generated by Meshy, the completed geometry, and the textured object, all rendered under two different poses.}
    \vspace{-4mm}
    \label{fig:result_combined}
\end{figure*}

\begin{figure}[ht]
    \centering
    \vspace{-2mm}
    \includegraphics[width=0.9\linewidth]{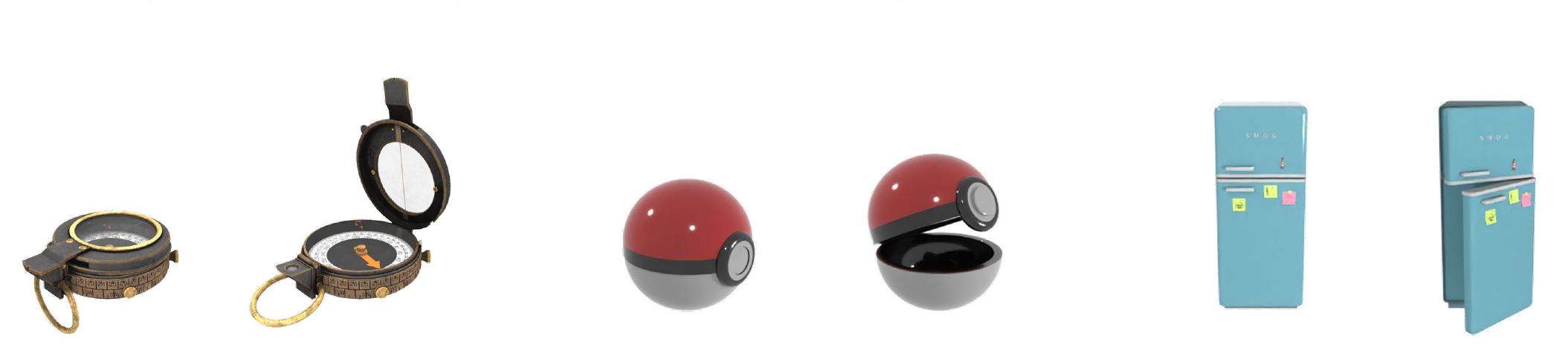}

    \caption{Results of \model~applied to hand-crafted meshes retrieved from Objaverse.}
     \vspace{-6mm}
    \label{fig:objaverse}
\end{figure}

\begin{figure*}[ht]
    \centering
    \includegraphics[width=0.9\linewidth]{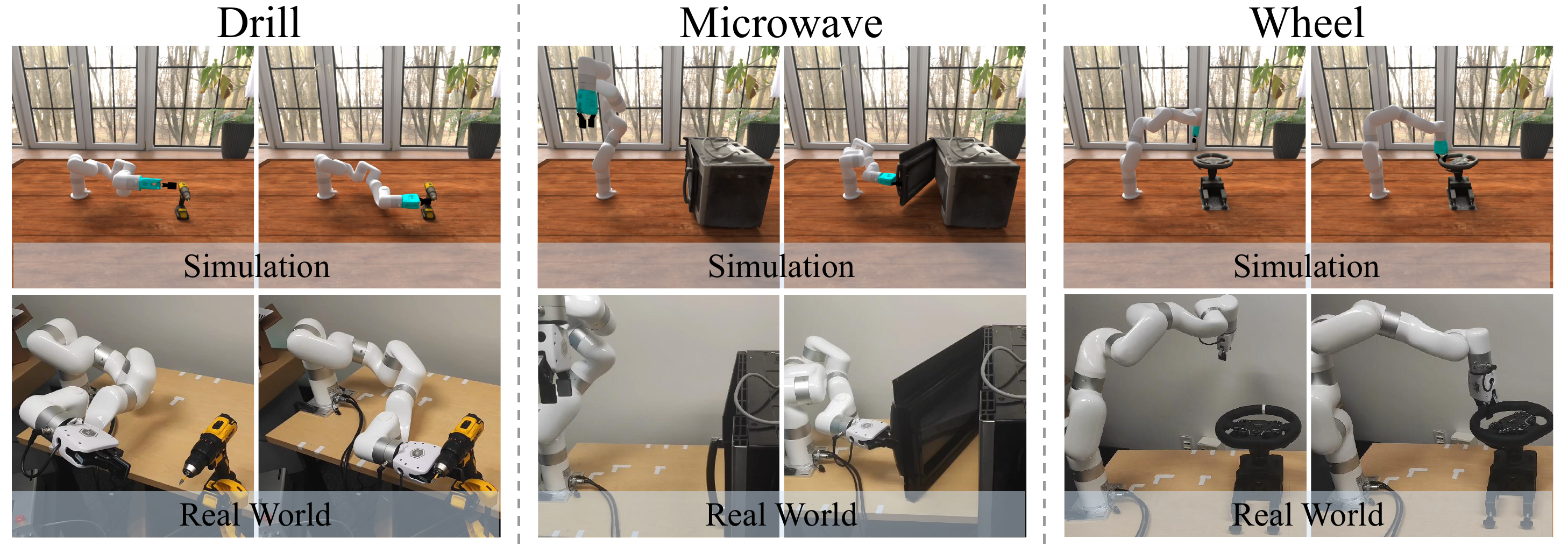}
    \vspace{-4mm}
    \caption{We build the digital twins of real-world objects in the simulation using our method. Then we sample trajectories completing the tasks in the simulation and replicate those trajectories in the real-world.}
    \label{fig:real}
\end{figure*}

\begin{table*}[t]
\centering
\begin{minipage}{0.48\textwidth}
\centering
\setlength{\tabcolsep}{3pt}
\vspace{-3mm}
\begin{tabular}{c|cc|cc}
\Xhline{2.5\arrayrulewidth}
 & \multicolumn{2}{c|}{ID} & \multicolumn{2}{c}{OOD} \\ \cline{2-5}
 & angle$\downarrow$ & pos$\downarrow$ & angle$\downarrow$ & pos$\downarrow$ \\ \hline

URDFormer       & 13.674 & 0.144 & 33.864 & 0.299 \\
Real2Code       & 7.950  & 0.167 & 24.035 & 0.194 \\
\textbf{Ours}   & \textbf{6.257} & \textbf{0.076} & \textbf{6.582} & \textbf{0.083} \\
\Xhline{2.5\arrayrulewidth}
\end{tabular}
\caption{Comparison of joint estimation accuracy—including angle and position error—between \model, URDFormer and Real2Code.}
\label{tab:former_real2code}
\end{minipage}
\hfill
\begin{minipage}{0.48\textwidth}
\centering
\setlength{\tabcolsep}{2pt}
\begin{tabular}{c|cc|cc}
\Xhline{2.5\arrayrulewidth}
& \multicolumn{2}{c|}{Laptop} & \multicolumn{2}{c}{Bucket} \\ \cline{2-5}
& original & aug & original & aug \\ \hline

succ. rate $\uparrow$ & 0.408 & \textbf{0.513} & 0.339 & \textbf{0.479} \\ \Xhline{2.5\arrayrulewidth}
\end{tabular}
\caption{Success rates of manipulation, comparing fine-tuning on original vs. augmented training sets. Evaluated on the augmented test set.}
\label{tab_policy}
\end{minipage}
\vspace{-6mm}
\end{table*}

\subsection{Applications}

\textbf{Articulated object generation}
By integrating an existing 3D generation model (Meshy\cite{Meshy}) to generate surface meshes as inputs to \model, our pipeline exploits the open-vocabulary 3D generation capabilities of the 3D generation model and inherits its generative paradigm. As in Figure \ref{fig:result_combined}, our extended pipeline generates high-quality articulated objects of various categories. 

\textbf{Annotate 3D object datasets}
Other than generating from scratch, \model can also start from existing artist designed meshes. For example, we annotate part segmentation and joint structures on objects from Objaverse \citep{deitke2024objaverse}. Such objects usually have high-quality mesh and textures and also with inner structures. Thus, we only execute the 3D segmentation part and articulation estimation part of our pipeline to annotate these objects. The results are demonstrated in Figure~\ref{fig:objaverse}. Objects in the teaser (Figure \ref{fig:teaser}) are also produced in this way.

\textbf{Real-to-Sim-to-Real}
In this experiment, we first reconstruct the 3D surface mesh of real-world objects using 2DGS \cite{huang20242d}. We then apply \model~to convert the reconstructed mesh into an articulated object represented in URDF format, making it compatible with simulation environments. Next, we sample action targets and use motion planning \citep{sucan2012open} to avoid collisions, generating a trajectory to successfully complete the task in simulation. Finally, we replicate the trajectory in the real world and observe that the robot arm can effectively execute the task.

We focus on three tasks involving different articulated objects: \textbf{pushing a drill trigger}, \textbf{opening a microwave door}, and \textbf{rotating a steering wheel}, as shown in Figure~\ref{fig:real}. Our results show that the generated trajectories can be successfully transferred to the real world, demonstrating minimal error in part segmentation and joint prediction.

\textbf{Policy learning in simulation}
We follow DexArt \cite{bao2023dexart} for articulated object manipulation policy learning and evaluate two tasks: opening a laptop lid and lifting a bucket. To augment the training set for the "bucket" and "laptop" experiments, we include additional articulated objects of the same categories generated by \model. Additionally, we diversify the testing set by incorporating some generated articulated objects. We fine-tune the checkpoints trained exclusively on the original dataset and evaluate their performance on the augmented testing set. The results, presented in Table~\ref{tab_policy}, demonstrate that the data generated by \model~effectively enhances robot learning. The articulated objects generated by \model are sourced from Objaverse. For visualizations of the data used and implementation details, please refer to Appendix Section \ref{sec_dataset}.

\section{Conclusion}
In this paper, we introduce $\model$, a novel framework for the critical task of converting open-vocabulary 3D meshes into their articulated counterparts. Our pipeline first segments 3D objects based on part-level semantics, estimates the articulation structure among object parts, and then refines the geometry and texture to repair flaws and enhance visual quality. By leveraging advancements in vision-language models and visual prompting techniques, we effectively segment parts and estimate articulation structures using common geometric cues. Our pipeline creates high-quality textured articulated objects from generated 3D meshes, hand-crafted 3D assets, and reconstructed meshes. The resulting assets can support the learning of articulated object manipulation skills in simulation, which can then be transferred to real-world robots.

\section{Limitations}
Although our method creates articulated objects with semantically correct parts, the articulation parameters are not always accurate or physically grounded. This limitation arises from three key challenges: (1) existing and generated 3D meshes often lack correct physical structures, (2) current 3D segmentation methods occasionally produce incorrect results, and (3) GPT4o exhibits biases in certain scenarios.
Additionally, some uncommon cases are not covered by our method—for example, when a drawer is designed to be pulled out up-front. 

If future advancements lead to the development of large VLMs with enhanced 3D reasoning capabilities, our pipeline could leverage these improvements to become more robust and reliable. Addressing these challenges to create fully accurate and physically grounded articulation parameters represents an exciting direction for future research.

\bibliography{example}  

\begin{thebibliography}{67}
\providecommand{\natexlab}[1]{#1}
\providecommand{\url}[1]{\texttt{#1}}
\expandafter\ifx\csname urlstyle\endcsname\relax
  \providecommand{\doi}[1]{doi: #1}\else
  \providecommand{\doi}{doi: \begingroup \urlstyle{rm}\Url}\fi

\bibitem[Deitke et~al.(2024)Deitke, Liu, Wallingford, Ngo, Michel, Kusupati, Fan, Laforte, Voleti, Gadre, et~al.]{deitke2024objaverse}
M.~Deitke, R.~Liu, M.~Wallingford, H.~Ngo, O.~Michel, A.~Kusupati, A.~Fan, C.~Laforte, V.~Voleti, S.~Y. Gadre, et~al.
\newblock Objaverse-xl: A universe of 10m+ 3d objects.
\newblock \emph{Advances in Neural Information Processing Systems}, 36, 2024.

\bibitem[Chen et~al.(2024)Chen, Walsman, Memmel, Mo, Fang, Vemuri, Wu, Fox, and Gupta]{chen2024urdformer}
Z.~Chen, A.~Walsman, M.~Memmel, K.~Mo, A.~Fang, K.~Vemuri, A.~Wu, D.~Fox, and A.~Gupta.
\newblock Urdformer: A pipeline for constructing articulated simulation environments from real-world images.
\newblock \emph{arXiv preprint arXiv:2405.11656}, 2024.

\bibitem[Yang et~al.(2024{\natexlab{a}})Yang, Jia, Zhi, and Huang]{yang2024physcene}
Y.~Yang, B.~Jia, P.~Zhi, and S.~Huang.
\newblock Physcene: Physically interactable 3d scene synthesis for embodied ai.
\newblock In \emph{Proceedings of the IEEE/CVF Conference on Computer Vision and Pattern Recognition}, pages 16262--16272, 2024{\natexlab{a}}.

\bibitem[Yang et~al.(2024{\natexlab{b}})Yang, Sun, Weihs, VanderBilt, Herrasti, Han, Wu, Haber, Krishna, Liu, et~al.]{yang2024holodeck}
Y.~Yang, F.-Y. Sun, L.~Weihs, E.~VanderBilt, A.~Herrasti, W.~Han, J.~Wu, N.~Haber, R.~Krishna, L.~Liu, et~al.
\newblock Holodeck: Language guided generation of 3d embodied ai environments.
\newblock In \emph{Proceedings of the IEEE/CVF Conference on Computer Vision and Pattern Recognition}, pages 16227--16237, 2024{\natexlab{b}}.

\bibitem[Wang et~al.(2024)Wang, Qiu, Liu, Chen, Cai, Wang, Wang, Xian, and Gan]{wang2024architect}
Y.~Wang, X.~Qiu, J.~Liu, Z.~Chen, J.~Cai, Y.~Wang, T.-H.~J. Wang, Z.~Xian, and C.~Gan.
\newblock Architect: Generating vivid and interactive 3d scenes with hierarchical 2d inpainting.
\newblock \emph{Advances in Neural Information Processing Systems}, 37:\penalty0 67575--67603, 2024.

\bibitem[Wang et~al.(2023)Wang, Xian, Chen, Wang, Wang, Fragkiadaki, Erickson, Held, and Gan]{wang2023robogen}
Y.~Wang, Z.~Xian, F.~Chen, T.-H. Wang, Y.~Wang, K.~Fragkiadaki, Z.~Erickson, D.~Held, and C.~Gan.
\newblock Robogen: Towards unleashing infinite data for automated robot learning via generative simulation.
\newblock \emph{arXiv preprint arXiv:2311.01455}, 2023.

\bibitem[Dalal et~al.(2023)Dalal, Mandlekar, Garrett, Handa, Salakhutdinov, and Fox]{dalal2023imitating}
M.~Dalal, A.~Mandlekar, C.~Garrett, A.~Handa, R.~Salakhutdinov, and D.~Fox.
\newblock Imitating task and motion planning with visuomotor transformers.
\newblock \emph{arXiv preprint arXiv:2305.16309}, 2023.

\bibitem[Ha et~al.(2023)Ha, Florence, and Song]{ha2023scaling}
H.~Ha, P.~Florence, and S.~Song.
\newblock Scaling up and distilling down: Language-guided robot skill acquisition.
\newblock In \emph{Conference on Robot Learning}, pages 3766--3777. PMLR, 2023.

\bibitem[Ma et~al.(2023)Ma, Liang, Wang, Huang, Bastani, Jayaraman, Zhu, Fan, and Anandkumar]{ma2023eureka}
Y.~J. Ma, W.~Liang, G.~Wang, D.-A. Huang, O.~Bastani, D.~Jayaraman, Y.~Zhu, L.~Fan, and A.~Anandkumar.
\newblock Eureka: Human-level reward design via coding large language models.
\newblock \emph{arXiv preprint arXiv:2310.12931}, 2023.

\bibitem[Long et~al.(2024)Long, Guo, Lin, Liu, Dou, Liu, Ma, Zhang, Habermann, Theobalt, et~al.]{long2024wonder3d}
X.~Long, Y.-C. Guo, C.~Lin, Y.~Liu, Z.~Dou, L.~Liu, Y.~Ma, S.-H. Zhang, M.~Habermann, C.~Theobalt, et~al.
\newblock Wonder3d: Single image to 3d using cross-domain diffusion.
\newblock In \emph{Proceedings of the IEEE/CVF Conference on Computer Vision and Pattern Recognition}, pages 9970--9980, 2024.

\bibitem[Hong et~al.(2023)Hong, Zhang, Gu, Bi, Zhou, Liu, Liu, Sunkavalli, Bui, and Tan]{hong2023lrm}
Y.~Hong, K.~Zhang, J.~Gu, S.~Bi, Y.~Zhou, D.~Liu, F.~Liu, K.~Sunkavalli, T.~Bui, and H.~Tan.
\newblock Lrm: Large reconstruction model for single image to 3d.
\newblock \emph{arXiv preprint arXiv:2311.04400}, 2023.

\bibitem[Xu et~al.(2024)Xu, Cheng, Gao, Wang, Gao, and Shan]{xu2024instantmesh}
J.~Xu, W.~Cheng, Y.~Gao, X.~Wang, S.~Gao, and Y.~Shan.
\newblock Instantmesh: Efficient 3d mesh generation from a single image with sparse-view large reconstruction models.
\newblock \emph{arXiv preprint arXiv:2404.07191}, 2024.

\bibitem[Shi et~al.(2023{\natexlab{a}})Shi, Chen, Zhang, Liu, Xu, Wei, Chen, Zeng, and Su]{shi2023zero123++}
R.~Shi, H.~Chen, Z.~Zhang, M.~Liu, C.~Xu, X.~Wei, L.~Chen, C.~Zeng, and H.~Su.
\newblock Zero123++: a single image to consistent multi-view diffusion base model.
\newblock \emph{arXiv preprint arXiv:2310.15110}, 2023{\natexlab{a}}.

\bibitem[Shi et~al.(2023{\natexlab{b}})Shi, Wang, Ye, Long, Li, and Yang]{shi2023mvdream}
Y.~Shi, P.~Wang, J.~Ye, M.~Long, K.~Li, and X.~Yang.
\newblock Mvdream: Multi-view diffusion for 3d generation.
\newblock \emph{arXiv preprint arXiv:2308.16512}, 2023{\natexlab{b}}.

\bibitem[Poole et~al.(2022)Poole, Jain, Barron, and Mildenhall]{poole2022dreamfusion}
B.~Poole, A.~Jain, J.~T. Barron, and B.~Mildenhall.
\newblock Dreamfusion: Text-to-3d using 2d diffusion.
\newblock \emph{arXiv preprint arXiv:2209.14988}, 2022.

\bibitem[Wang et~al.(2024)Wang, Lu, Wang, Bao, Li, Su, and Zhu]{wang2024prolificdreamer}
Z.~Wang, C.~Lu, Y.~Wang, F.~Bao, C.~Li, H.~Su, and J.~Zhu.
\newblock Prolificdreamer: High-fidelity and diverse text-to-3d generation with variational score distillation.
\newblock \emph{Advances in Neural Information Processing Systems}, 36, 2024.

\bibitem[Qiu et~al.(2024)Qiu, Chen, Gu, Zuo, Xu, Wu, Yuan, Dong, Bo, and Han]{qiu2024richdreamer}
L.~Qiu, G.~Chen, X.~Gu, Q.~Zuo, M.~Xu, Y.~Wu, W.~Yuan, Z.~Dong, L.~Bo, and X.~Han.
\newblock Richdreamer: A generalizable normal-depth diffusion model for detail richness in text-to-3d.
\newblock In \emph{Proceedings of the IEEE/CVF Conference on Computer Vision and Pattern Recognition}, pages 9914--9925, 2024.

\bibitem[Gao et~al.(2019)Gao, Yang, Wu, Yuan, Fu, Lai, and Zhang]{gao2019sdm}
L.~Gao, J.~Yang, T.~Wu, Y.-J. Yuan, H.~Fu, Y.-K. Lai, and H.~Zhang.
\newblock Sdm-net: Deep generative network for structured deformable mesh.
\newblock \emph{ACM Transactions on Graphics (TOG)}, 38\penalty0 (6):\penalty0 1--15, 2019.

\bibitem[Yang et~al.(2022)Yang, Mo, Lai, Guibas, and Gao]{yang2022dsg}
J.~Yang, K.~Mo, Y.-K. Lai, L.~J. Guibas, and L.~Gao.
\newblock Dsg-net: Learning disentangled structure and geometry for 3d shape generation.
\newblock \emph{ACM Transactions on Graphics (TOG)}, 42\penalty0 (1):\penalty0 1--17, 2022.

\bibitem[Mo et~al.(2019)Mo, Guerrero, Yi, Su, Wonka, Mitra, and Guibas]{mo2019structurenet}
K.~Mo, P.~Guerrero, L.~Yi, H.~Su, P.~Wonka, N.~Mitra, and L.~J. Guibas.
\newblock Structurenet: Hierarchical graph networks for 3d shape generation.
\newblock \emph{arXiv preprint arXiv:1908.00575}, 2019.

\bibitem[Wu et~al.(2020)Wu, Zhuang, Xu, Zhang, and Chen]{wu2020pq}
R.~Wu, Y.~Zhuang, K.~Xu, H.~Zhang, and B.~Chen.
\newblock Pq-net: A generative part seq2seq network for 3d shapes.
\newblock In \emph{Proceedings of the IEEE/CVF Conference on Computer Vision and Pattern Recognition}, pages 829--838, 2020.

\bibitem[Nakayama et~al.(2023)Nakayama, Uy, Huang, Hu, Li, and Guibas]{nakayama2023difffacto}
G.~K. Nakayama, M.~A. Uy, J.~Huang, S.-M. Hu, K.~Li, and L.~Guibas.
\newblock Difffacto: Controllable part-based 3d point cloud generation with cross diffusion.
\newblock In \emph{Proceedings of the IEEE/CVF International Conference on Computer Vision}, pages 14257--14267, 2023.

\bibitem[Koo et~al.(2023)Koo, Yoo, Nguyen, and Sung]{koo2023salad}
J.~Koo, S.~Yoo, M.~H. Nguyen, and M.~Sung.
\newblock Salad: Part-level latent diffusion for 3d shape generation and manipulation.
\newblock In \emph{Proceedings of the IEEE/CVF International Conference on Computer Vision}, pages 14441--14451, 2023.

\bibitem[Liu et~al.(2024)Liu, Lin, Liu, Long, Dou, Guo, Luo, and Wang]{liu2024part123}
A.~Liu, C.~Lin, Y.~Liu, X.~Long, Z.~Dou, H.-X. Guo, P.~Luo, and W.~Wang.
\newblock Part123: Part-aware 3d reconstruction from a single-view image.
\newblock In \emph{ACM SIGGRAPH 2024 Conference Papers}, pages 1--12, 2024.

\bibitem[Jiang et~al.(2022)Jiang, Hsu, and Zhu]{jiang2022ditto}
Z.~Jiang, C.-C. Hsu, and Y.~Zhu.
\newblock Ditto: Building digital twins of articulated objects from interaction.
\newblock In \emph{Proceedings of the IEEE/CVF Conference on Computer Vision and Pattern Recognition}, pages 5616--5626, 2022.

\bibitem[Liu et~al.(2023)Liu, Mahdavi-Amiri, and Savva]{liu2023paris}
J.~Liu, A.~Mahdavi-Amiri, and M.~Savva.
\newblock Paris: Part-level reconstruction and motion analysis for articulated objects.
\newblock In \emph{Proceedings of the IEEE/CVF International Conference on Computer Vision}, pages 352--363, 2023.

\bibitem[Lei et~al.(2023)Lei, Deng, Shen, Guibas, and Daniilidis]{lei2023nap}
J.~Lei, C.~Deng, W.~B. Shen, L.~J. Guibas, and K.~Daniilidis.
\newblock Nap: Neural 3d articulated object prior.
\newblock \emph{Advances in Neural Information Processing Systems}, 36:\penalty0 31878--31894, 2023.

\bibitem[Liu et~al.(2024)Liu, Tam, Mahdavi-Amiri, and Savva]{liu2024cage}
J.~Liu, H.~I.~I. Tam, A.~Mahdavi-Amiri, and M.~Savva.
\newblock Cage: Controllable articulation generation.
\newblock In \emph{Proceedings of the IEEE/CVF Conference on Computer Vision and Pattern Recognition}, pages 17880--17889, 2024.

\bibitem[Mandi et~al.(2024)Mandi, Weng, Bauer, and Song]{mandi2024real2code}
Z.~Mandi, Y.~Weng, D.~Bauer, and S.~Song.
\newblock Real2code: Reconstruct articulated objects via code generation.
\newblock \emph{arXiv preprint arXiv:2406.08474}, 2024.

\bibitem[Nie et~al.(2023)Nie, Gadre, Ehsani, and Song]{nie2023structure}
N.~Nie, S.~Y. Gadre, K.~Ehsani, and S.~Song.
\newblock Structure from action: Learning interactions for 3d articulated object structure discovery.
\newblock In \emph{2023 IEEE/RSJ International Conference on Intelligent Robots and Systems (IROS)}, pages 1222--1229. IEEE, 2023.

\bibitem[Gadre et~al.(2021)Gadre, Ehsani, and Song]{gadre2021act}
S.~Y. Gadre, K.~Ehsani, and S.~Song.
\newblock Act the part: Learning interaction strategies for articulated object part discovery.
\newblock In \emph{Proceedings of the IEEE/CVF International Conference on Computer Vision}, pages 15752--15761, 2021.

\bibitem[Huang et~al.(2021)Huang, Wang, Birdal, Sung, Arrigoni, Hu, and Guibas]{huang2021multibodysync}
J.~Huang, H.~Wang, T.~Birdal, M.~Sung, F.~Arrigoni, S.-M. Hu, and L.~J. Guibas.
\newblock Multibodysync: Multi-body segmentation and motion estimation via 3d scan synchronization.
\newblock In \emph{Proceedings of the IEEE/CVF Conference on Computer Vision and Pattern Recognition}, pages 7108--7118, 2021.

\bibitem[Xiang et~al.(2020)Xiang, Qin, Mo, Xia, Zhu, Liu, Liu, Jiang, Yuan, Wang, et~al.]{xiang2020sapien}
F.~Xiang, Y.~Qin, K.~Mo, Y.~Xia, H.~Zhu, F.~Liu, M.~Liu, H.~Jiang, Y.~Yuan, H.~Wang, et~al.
\newblock Sapien: A simulated part-based interactive environment.
\newblock In \emph{Proceedings of the IEEE/CVF conference on computer vision and pattern recognition}, pages 11097--11107, 2020.

\bibitem[Wang et~al.(2019)Wang, Zhou, Shi, Chen, Zhao, and Xu]{wang2019shape2motion}
X.~Wang, B.~Zhou, Y.~Shi, X.~Chen, Q.~Zhao, and K.~Xu.
\newblock Shape2motion: Joint analysis of motion parts and attributes from 3d shapes.
\newblock In \emph{Proceedings of the IEEE/CVF Conference on Computer Vision and Pattern Recognition}, pages 8876--8884, 2019.

\bibitem[Geng et~al.(2023)Geng, Xu, Zhao, Xu, Yi, Huang, and Wang]{geng2023gapartnet}
H.~Geng, H.~Xu, C.~Zhao, C.~Xu, L.~Yi, S.~Huang, and H.~Wang.
\newblock Gapartnet: Cross-category domain-generalizable object perception and manipulation via generalizable and actionable parts.
\newblock In \emph{Proceedings of the IEEE/CVF Conference on Computer Vision and Pattern Recognition}, pages 7081--7091, 2023.

\bibitem[Achiam et~al.(2023)Achiam, Adler, Agarwal, Ahmad, Akkaya, Aleman, Almeida, Altenschmidt, Altman, Anadkat, et~al.]{achiam2023gpt}
J.~Achiam, S.~Adler, S.~Agarwal, L.~Ahmad, I.~Akkaya, F.~L. Aleman, D.~Almeida, J.~Altenschmidt, S.~Altman, S.~Anadkat, et~al.
\newblock Gpt-4 technical report.
\newblock \emph{arXiv preprint arXiv:2303.08774}, 2023.

\bibitem[Kirillov et~al.(2023)Kirillov, Mintun, Ravi, Mao, Rolland, Gustafson, Xiao, Whitehead, Berg, Lo, et~al.]{kirillov2023segment}
A.~Kirillov, E.~Mintun, N.~Ravi, H.~Mao, C.~Rolland, L.~Gustafson, T.~Xiao, S.~Whitehead, A.~C. Berg, W.-Y. Lo, et~al.
\newblock Segment anything.
\newblock In \emph{Proceedings of the IEEE/CVF International Conference on Computer Vision}, pages 4015--4026, 2023.

\bibitem[Rombach et~al.(2022)Rombach, Blattmann, Lorenz, Esser, and Ommer]{rombach2022high}
R.~Rombach, A.~Blattmann, D.~Lorenz, P.~Esser, and B.~Ommer.
\newblock High-resolution image synthesis with latent diffusion models.
\newblock In \emph{Proceedings of the IEEE/CVF conference on computer vision and pattern recognition}, pages 10684--10695, 2022.

\bibitem[Liu et~al.(2023)Liu, Zhu, Cai, Han, Ling, Porikli, and Su]{liu2023partslip}
M.~Liu, Y.~Zhu, H.~Cai, S.~Han, Z.~Ling, F.~Porikli, and H.~Su.
\newblock Partslip: Low-shot part segmentation for 3d point clouds via pretrained image-language models.
\newblock In \emph{Proceedings of the IEEE/CVF conference on computer vision and pattern recognition}, pages 21736--21746, 2023.

\bibitem[Zhou et~al.(2023)Zhou, Gu, Li, Liu, Fang, and Su]{zhou2023partslip++}
Y.~Zhou, J.~Gu, X.~Li, M.~Liu, Y.~Fang, and H.~Su.
\newblock Partslip++: Enhancing low-shot 3d part segmentation via multi-view instance segmentation and maximum likelihood estimation.
\newblock \emph{arXiv preprint arXiv:2312.03015}, 2023.

\bibitem[Umam et~al.(2024)Umam, Yang, Chen, Chuang, and Lin]{umam2024partdistill}
A.~Umam, C.-K. Yang, M.-H. Chen, J.-H. Chuang, and Y.-Y. Lin.
\newblock Partdistill: 3d shape part segmentation by vision-language model distillation.
\newblock In \emph{Proceedings of the IEEE/CVF Conference on Computer Vision and Pattern Recognition}, pages 3470--3479, 2024.

\bibitem[Xue et~al.(2023)Xue, Chen, Liu, and Sun]{xue2023zerops}
Y.~Xue, N.~Chen, J.~Liu, and W.~Sun.
\newblock Zerops: High-quality cross-modal knowledge transfer for zero-shot 3d part segmentation.
\newblock \emph{arXiv preprint arXiv:2311.14262}, 2023.

\bibitem[Yang et~al.(2024)Yang, Huang, Guo, Lu, Wu, Lam, Cao, and Liu]{yang2024sampart3d}
Y.~Yang, Y.~Huang, Y.-C. Guo, L.~Lu, X.~Wu, E.~Y. Lam, Y.-P. Cao, and X.~Liu.
\newblock Sampart3d: Segment any part in 3d objects.
\newblock \emph{arXiv preprint arXiv:2411.07184}, 2024.

\bibitem[Li et~al.(2020)Li, Wang, Yi, Guibas, Abbott, and Song]{li2020category}
X.~Li, H.~Wang, L.~Yi, L.~J. Guibas, A.~L. Abbott, and S.~Song.
\newblock Category-level articulated object pose estimation.
\newblock In \emph{Proceedings of the IEEE/CVF conference on computer vision and pattern recognition}, pages 3706--3715, 2020.

\bibitem[Zeng et~al.(2024)Zeng, Zhang, Wu, Wang, Ye, and Li]{zeng2024mars}
H.~Zeng, P.~Zhang, C.~Wu, J.~Wang, T.~Ye, and F.~Li.
\newblock Mars: Multimodal active robotic sensing for articulated characterization.
\newblock \emph{arXiv preprint arXiv:2407.01191}, 2024.

\bibitem[Yang et~al.(2025)Yang, Guo, Huang, Zou, Yu, Li, Cao, and Liu]{yang2025holopart}
Y.~Yang, Y.-C. Guo, Y.~Huang, Z.-X. Zou, Z.~Yu, Y.~Li, Y.-P. Cao, and X.~Liu.
\newblock Holopart: Generative 3d part amodal segmentation.
\newblock \emph{arXiv preprint arXiv:2504.07943}, 2025.

\bibitem[Mes(2025)]{Meshy}
Meshy ai, 2025.
\newblock URL \url{https://www.meshy.ai}.

\bibitem[Hsu et~al.(2023)Hsu, Jiang, and Zhu]{hsu2023ditto}
C.-C. Hsu, Z.~Jiang, and Y.~Zhu.
\newblock Ditto in the house: Building articulation models of indoor scenes through interactive perception.
\newblock In \emph{2023 IEEE International Conference on Robotics and Automation (ICRA)}, pages 3933--3939. IEEE, 2023.

\bibitem[Wang et~al.(2022)Wang, Wu, Mo, Ke, Fan, Guibas, and Dong]{wang2022adaafford}
Y.~Wang, R.~Wu, K.~Mo, J.~Ke, Q.~Fan, L.~J. Guibas, and H.~Dong.
\newblock Adaafford: Learning to adapt manipulation affordance for 3d articulated objects via few-shot interactions.
\newblock In \emph{European conference on computer vision}, pages 90--107. Springer, 2022.

\bibitem[Weng et~al.(2024)Weng, Wen, Tremblay, Blukis, Fox, Guibas, and Birchfield]{weng2024neural}
Y.~Weng, B.~Wen, J.~Tremblay, V.~Blukis, D.~Fox, L.~Guibas, and S.~Birchfield.
\newblock Neural implicit representation for building digital twins of unknown articulated objects.
\newblock In \emph{Proceedings of the IEEE/CVF Conference on Computer Vision and Pattern Recognition}, pages 3141--3150, 2024.

\bibitem[Wei et~al.(2022)Wei, Chabra, Ma, Lassner, Zollh{\"o}fer, Rusinkiewicz, Sweeney, Newcombe, and Slavcheva]{wei2022self}
F.~Wei, R.~Chabra, L.~Ma, C.~Lassner, M.~Zollh{\"o}fer, S.~Rusinkiewicz, C.~Sweeney, R.~Newcombe, and M.~Slavcheva.
\newblock Self-supervised neural articulated shape and appearance models.
\newblock In \emph{Proceedings of the IEEE/CVF Conference on Computer Vision and Pattern Recognition}, pages 15816--15826, 2022.

\bibitem[Jiang et~al.(2022)Jiang, Mao, Savva, and Chang]{jiang2022opd}
H.~Jiang, Y.~Mao, M.~Savva, and A.~X. Chang.
\newblock Opd: Single-view 3d openable part detection.
\newblock In \emph{European Conference on Computer Vision}, pages 410--426. Springer, 2022.

\bibitem[Sun et~al.(2024)Sun, Jiang, Savva, and Chang]{sun2024opdmulti}
X.~Sun, H.~Jiang, M.~Savva, and A.~Chang.
\newblock Opdmulti: Openable part detection for multiple objects.
\newblock In \emph{2024 International Conference on 3D Vision (3DV)}, pages 169--178. IEEE, 2024.

\bibitem[Yan et~al.(2020)Yan, Hu, Yan, Chen, Van~Kaick, Zhang, and Huang]{yan2020rpm}
Z.~Yan, R.~Hu, X.~Yan, L.~Chen, O.~Van~Kaick, H.~Zhang, and H.~Huang.
\newblock Rpm-net: recurrent prediction of motion and parts from point cloud.
\newblock \emph{arXiv preprint arXiv:2006.14865}, 2020.

\bibitem[Liu et~al.(2022)Liu, Xue, Xu, Fu, and Lu]{liu2022toward}
L.~Liu, H.~Xue, W.~Xu, H.~Fu, and C.~Lu.
\newblock Toward real-world category-level articulation pose estimation.
\newblock \emph{IEEE Transactions on Image Processing}, 31:\penalty0 1072--1083, 2022.

\bibitem[Liu et~al.(2023)Liu, Zhang, Hu, Huang, Wang, and Yi]{liu2023self}
X.~Liu, J.~Zhang, R.~Hu, H.~Huang, H.~Wang, and L.~Yi.
\newblock Self-supervised category-level articulated object pose estimation with part-level se (3) equivariance.
\newblock \emph{arXiv preprint arXiv:2302.14268}, 2023.

\bibitem[Liu et~al.(2024{\natexlab{a}})Liu, Iliash, Chang, Savva, and Mahdavi-Amiri]{liu2024singapo}
J.~Liu, D.~Iliash, A.~X. Chang, M.~Savva, and A.~Mahdavi-Amiri.
\newblock Singapo: Single image controlled generation of articulated parts in objects.
\newblock \emph{arXiv preprint arXiv:2410.16499}, 2024{\natexlab{a}}.

\bibitem[Liu et~al.(2024{\natexlab{b}})Liu, Savva, and Mahdavi-Amiri]{liu2024survey}
J.~Liu, M.~Savva, and A.~Mahdavi-Amiri.
\newblock Survey on modeling of articulated objects.
\newblock \emph{arXiv preprint arXiv:2403.14937}, 2024{\natexlab{b}}.

\bibitem[Chen et~al.(2024)Chen, Shapovalov, Laina, Monnier, Wang, Novotny, and Vedaldi]{chen2024partgen}
M.~Chen, R.~Shapovalov, I.~Laina, T.~Monnier, J.~Wang, D.~Novotny, and A.~Vedaldi.
\newblock Partgen: Part-level 3d generation and reconstruction with multi-view diffusion models.
\newblock \emph{arXiv preprint arXiv:2412.18608}, 2024.

\bibitem[Ren et~al.(2024)Ren, Chen, Jiang, Zeng, Xiong, Liu, Ma, Shen, Gao, Jiang, et~al.]{ren2024dino}
T.~Ren, Y.~Chen, Q.~Jiang, Z.~Zeng, Y.~Xiong, W.~Liu, Z.~Ma, J.~Shen, Y.~Gao, X.~Jiang, et~al.
\newblock Dino-x: A unified vision model for open-world object detection and understanding.
\newblock \emph{arXiv preprint arXiv:2411.14347}, 2024.

\bibitem[Li et~al.(2022)Li, Zhang, Zhang, Yang, Li, Zhong, Wang, Yuan, Zhang, Hwang, et~al.]{li2022grounded}
L.~H. Li, P.~Zhang, H.~Zhang, J.~Yang, C.~Li, Y.~Zhong, L.~Wang, L.~Yuan, L.~Zhang, J.-N. Hwang, et~al.
\newblock Grounded language-image pre-training.
\newblock In \emph{Proceedings of the IEEE/CVF Conference on Computer Vision and Pattern Recognition}, pages 10965--10975, 2022.

\bibitem[Landrieu and Obozinski(2017)]{landrieu2017cut}
L.~Landrieu and G.~Obozinski.
\newblock Cut pursuit: Fast algorithms to learn piecewise constant functions on general weighted graphs.
\newblock \emph{SIAM Journal on Imaging Sciences}, 10\penalty0 (4):\penalty0 1724--1766, 2017.

\bibitem[Authors(2024)]{Genesis}
G.~Authors.
\newblock Genesis: A universal and generative physics engine for robotics and beyond, December 2024.
\newblock URL \url{https://github.com/Genesis-Embodied-AI/Genesis}.

\bibitem[Yang et~al.(2023)Yang, Zhang, Li, Zou, Li, and Gao]{yang2023set}
J.~Yang, H.~Zhang, F.~Li, X.~Zou, C.~Li, and J.~Gao.
\newblock Set-of-mark prompting unleashes extraordinary visual grounding in gpt-4v.
\newblock \emph{arXiv preprint arXiv:2310.11441}, 2023.

\bibitem[Huang et~al.(2024)Huang, Yu, Chen, Geiger, and Gao]{huang20242d}
B.~Huang, Z.~Yu, A.~Chen, A.~Geiger, and S.~Gao.
\newblock 2d gaussian splatting for geometrically accurate radiance fields.
\newblock In \emph{ACM SIGGRAPH 2024 conference papers}, pages 1--11, 2024.

\bibitem[Sucan et~al.(2012)Sucan, Moll, and Kavraki]{sucan2012open}
I.~A. Sucan, M.~Moll, and L.~E. Kavraki.
\newblock The open motion planning library.
\newblock \emph{IEEE Robotics \& Automation Magazine}, 19\penalty0 (4):\penalty0 72--82, 2012.

\bibitem[Bao et~al.(2023)Bao, Xu, Qin, and Wang]{bao2023dexart}
C.~Bao, H.~Xu, Y.~Qin, and X.~Wang.
\newblock Dexart: Benchmarking generalizable dexterous manipulation with articulated objects.
\newblock In \emph{Proceedings of the IEEE/CVF Conference on Computer Vision and Pattern Recognition}, pages 21190--21200, 2023.

\end{thebibliography}

\newpage
\appendix
\onecolumn
\section{More quantitative results on joint parameter estimation}
\label{sec_quanti}
In this section, we provide additional comparisons with methods not covered in the main paper for joint parameter estimation. Specifically, we include two generative approaches—NAP and CAGE—that leverage diffusion models to synthesize articulated objects and can generate joint configurations conditioned on object parts. In addition to these generative baselines, we also compare against discriminative methods such as ANCSH\cite{li2020category}, OPD\cite{jiang2022opd}, and OPDmulti\cite{sun2024opdmulti}. These methods take a single observation (e.g., RGB, depth, or both) as input and jointly predict part segmentation and articulation parameters. Together, these methods provide a broader context for evaluating joint estimation performance.

We first evaluate performance in the single-observation setting, using the OPD-Synth dataset. In this experiment, we compare \model against ANCSH, OPD, and OPDmulti for articulation parameter estimation from a single-view input. ANCSH uses a single-view point cloud as input, while OPD and OPDmulti operate on RGB images, optionally supplemented with depth. All three methods jointly predict part segmentation and joint parameters. To ensure a fair comparison, we adapt \model to this setting by restricting segmentation to a single image, which is then projected into a point cloud with part labels for articulation parameter estimation. As reported in Table~\ref{tab:single_obs}, \model~consistently outperforms all three methods within their respective training domains, despite being designed as a training-free, open-vocabulary approach.

We also evaluate articulation parameter estimation using ground-truth shapes of object parts as input, from PartNet-Mobility. Here, we compare \model with the generative diffusion-based models NAP and CAGE, both of which represent articulated objects as part graphs. NAP predicts joint parameters conditioned on part-level geometry descriptors such as bounding boxes, spatial locations, and learned shape latents. CAGE, by contrast, generates joint axes and ranges given attributes including part bounding boxes, semantic labels, and joint types. Following the experimental setup in CAGE, we use the PartNet-Mobility dataset and retrain NAP on the same train-test split. We evaluate both in-domain performance on test categories and generalization to held-out categories. As shown in Table\ref{tab:part_mesh}, while all three methods perform comparably in-domain, \model~substantially outperforms NAP and CAGE on previously unseen object categories, demonstrating superior generalization.

\begin{table}[h]
\begin{center}
\begin{tabular}{ccccc}
\Xhline{2.5\arrayrulewidth}
\multicolumn{1}{c|}{\textbf{}}                & ANCSH       & OPD         & OPDmulti   & \textbf{Ours}       \\ \cline{1-5}
\multicolumn{1}{c|}{angle error $\downarrow$} & 6.74                 & 10.73                & 9.66                 & \textbf{5.37}   \\
\multicolumn{1}{c|}{position error $\downarrow$}  & 0.065                & 0.117                & 0.108                & \textbf{0.049}   \\ \Xhline{2.5\arrayrulewidth}
\multicolumn{1}{l}{}                          & \multicolumn{1}{l}{} & \multicolumn{1}{l}{} & \multicolumn{1}{l}{} & \multicolumn{1}{l}{}

\end{tabular}

\caption{The results of single observation joint estimation. The dataset for training and testing is the one-door dataset (a subset of OPDsynth), which is used to train ANCSH as provided in its PyTorch version Github repository.}

\label{tab:single_obs}
\end{center}
\end{table}

\begin{table}[h]
\begin{center}
\begin{tabular}{c|cccc}
  \Xhline{2.5\arrayrulewidth}
     & \multicolumn{2}{c|}{angle error}                   & \multicolumn{2}{c}{position error}       \\ \cline{2-5} 
     & \multicolumn{1}{c}{ID $\downarrow$} & \multicolumn{1}{c|}{OOD $\downarrow$} & \multicolumn{1}{c}{ID $\downarrow$} & {OOD $\downarrow$}           \\ \hline
NAP  & 31.89                   & 42.23                    & 0.53                    & 0.225          \\
CAGE & 8.96                    & 58.64                    & \textbf{0.136}          & 0.192          \\
\textbf{Ours} & \textbf{7.90}           & \textbf{4.81}            & 0.190                   & \textbf{0.075} \\ \Xhline{2.5\arrayrulewidth}
\end{tabular}
\vspace{6pt}
\caption{The results for object part mesh joint estimation are evaluated on two sets of testing object categories: \textbf{in-domain (ID)}, consisting of categories included in CAGE's training set, and \textbf{out-of-domain (OOD)}, consisting of some categories from PartNet-Mobility that are not part of CAGE's training set.}
\label{tab:part_mesh}
\end{center}
\end{table}

\section{Ablation study of VLM}
\label{sec_abla_vlm}
\begin{table}[h]
\centering
\begin{tabular}{c|cc|cc}
\Xhline{2.5\arrayrulewidth}
 & \multicolumn{2}{c|}{ID} & \multicolumn{2}{c}{OOD} \\ \cline{2-5}
 & angle$\downarrow$ & pos$\downarrow$ & angle$\downarrow$ & pos$\downarrow$ \\ \hline
URDFormer         & 13.674 & 0.144 & 33.864 & 0.299 \\
Real2Code         & 7.950  & 0.167 & 24.035 & 0.194 \\
Ours (Claude)     & 9.291  & 0.160 & 7.541  & 0.096 \\
Ours (Gemini)     & 6.477  & 0.239 & 6.612  & 0.081 \\
\textbf{Ours (GPT)}     & \textbf{6.257} & \textbf{0.076} & \textbf{6.582} & \textbf{0.083} \\
\Xhline{2.5\arrayrulewidth}
\end{tabular}
\vspace{3pt}
\caption{Comparison of joint estimation accuracy (angle and position error) using different vision-language models.}
\label{tab_abla_vlm}
\end{table}

To investigate the effect of different vision-language model (VLM) choices, we conducted joint parameter estimation experiments using three VLMs: GPT-4o, Claude 3.7 Sonnet, and Gemini 2.0 Flash. Ground-truth part meshes from selected PartNet-Mobility categories (consistent with the setting used in the comparison with URDFormer and Real2Code, Table~\ref{tab:former_real2code}) were used as input for part segmentation and joint prediction. The results, presented in Table~\ref{tab_abla_vlm}, indicate that GPT-4o slightly outperforms Claude and Gemini, although the performance gap is relatively small. Notably, all three variants significantly outperform URDFormer and Real2Code across both in-domain and out-of-domain settings.

\section{Implementation of policy learning experiment}
\label{sec_dataset}
In this experiment, we evaluate two tasks proposed by DexArt \cite{bao2023dexart}. The first task is lifting a bucket, where a dexterous hand must grasp the bucket handle and lift it. In the original implementation, the bucket handle is thickened to prevent severe penetration, and we adhere to this setting. The second task is opening a laptop lid, where a dexterous hand must open the lid to a 90-degree angle.

\begin{figure*}[htbp]
    \centering
    \includegraphics[width=0.9\linewidth]{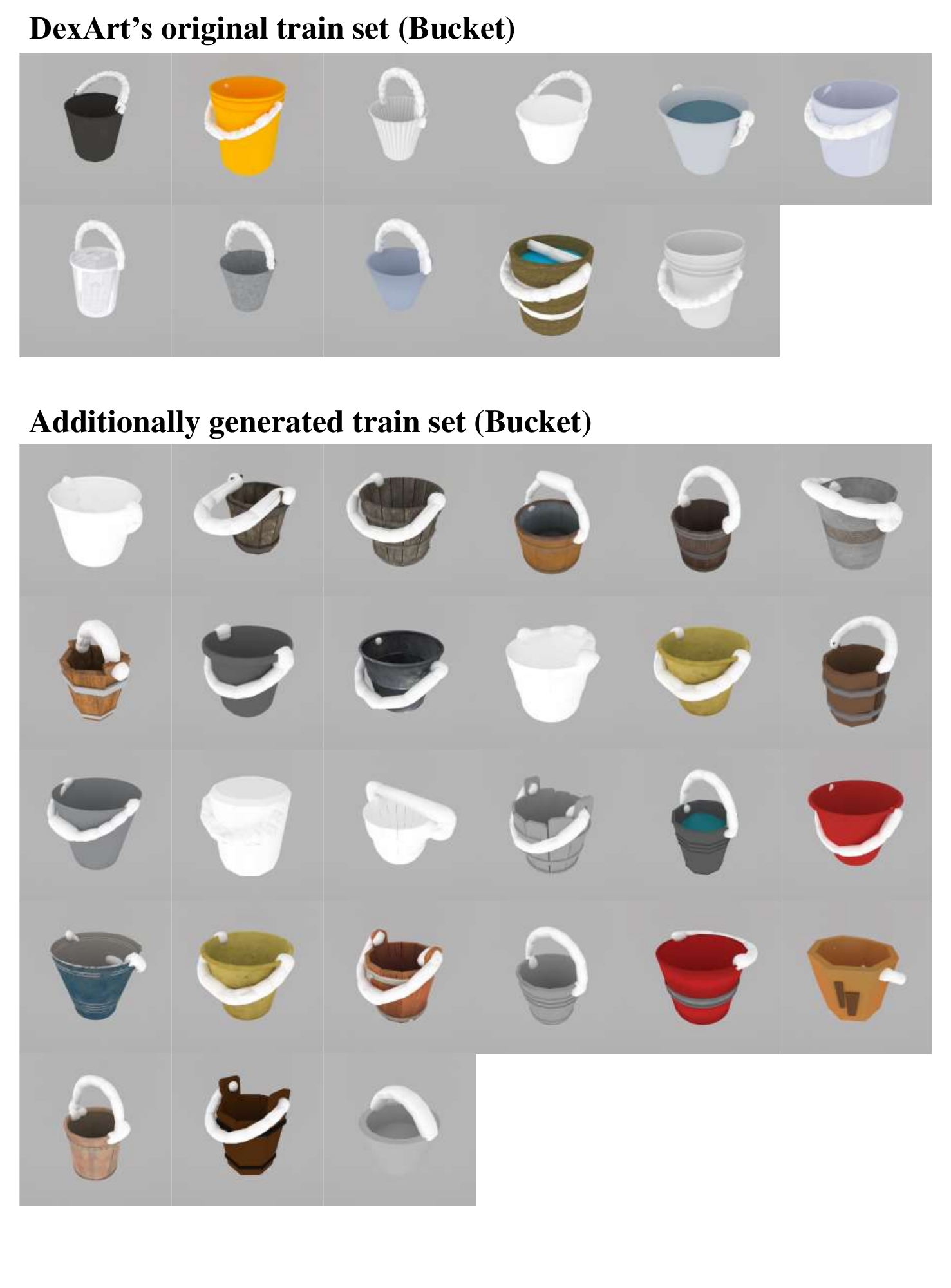}
    \vspace{-4mm}
    \caption{Visualization of the original training set for the bucket task from DexArt \cite{bao2023dexart} and the augmented training set collected using \model.}
    \vspace{-4mm}
    \label{fig:bk_train}
\end{figure*}

\begin{figure*}[htbp]
    \centering
    \includegraphics[width=0.9\linewidth]{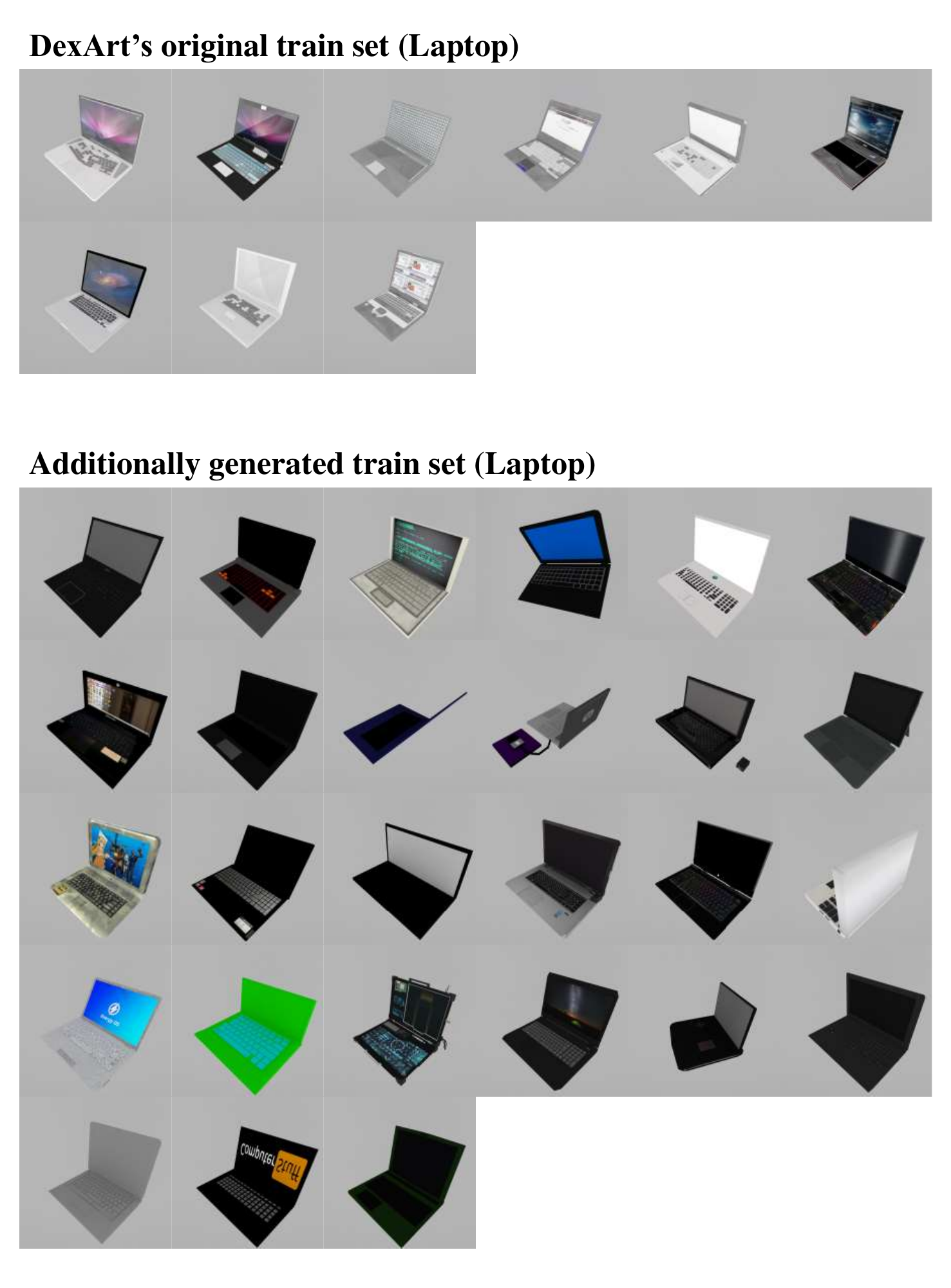}
    \vspace{-4mm}
    \caption{Visualization of the original training set for the laptop task from DexArt \cite{bao2023dexart} and the augmented training set collected using \model.}
    \vspace{-4mm}
    \label{fig:lp_train}
\end{figure*}

\begin{figure*}[htbp]
    \centering
    \includegraphics[width=0.9\linewidth]{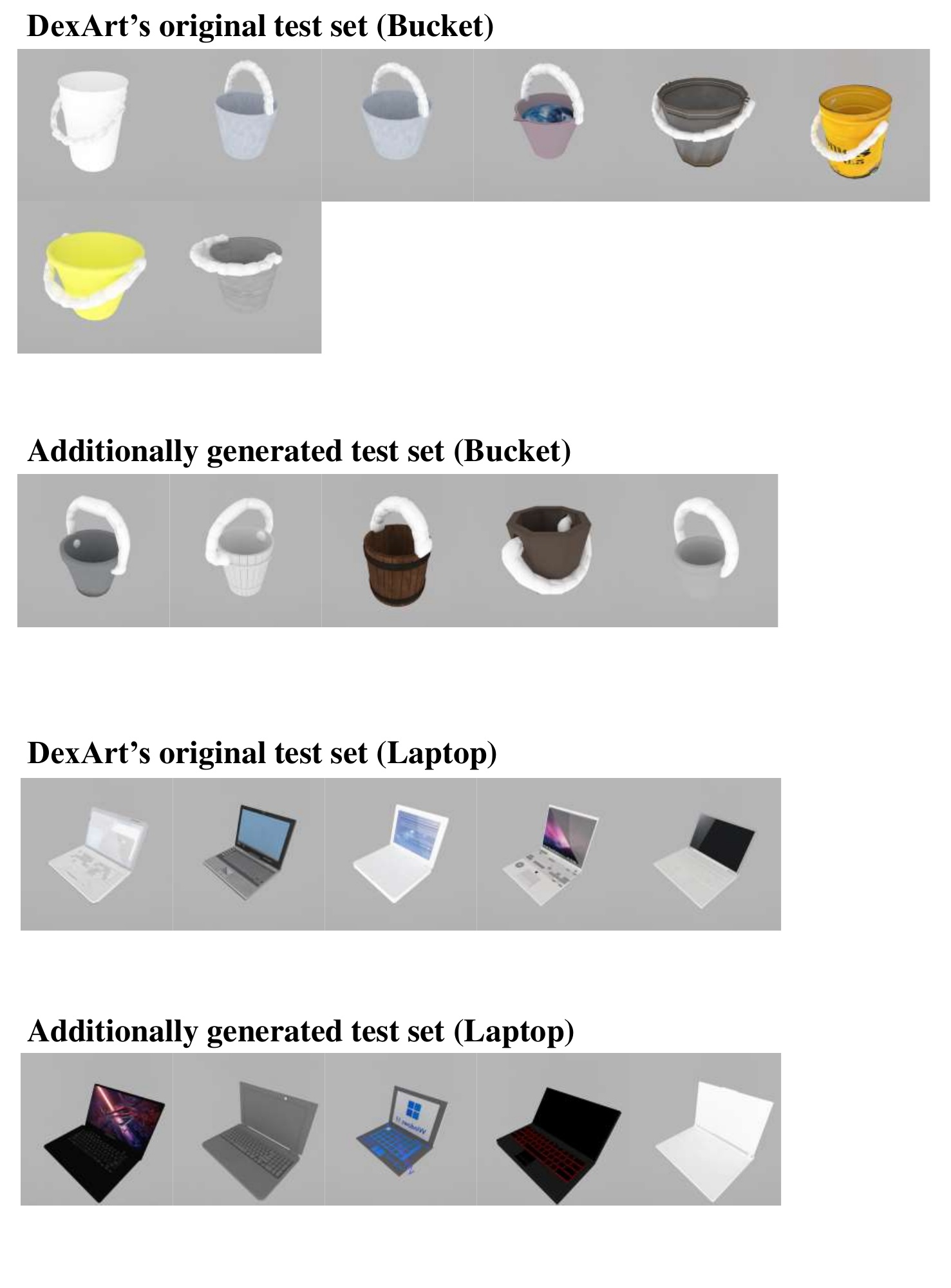}
    \vspace{-4mm}
    \caption{Visualization of the original test set used by DexArt \cite{bao2023dexart} and augmented test set collected using \model.}
    \vspace{-4mm}
    \label{fig:ltest}
\end{figure*}

For both tasks, we fine-tune the "no-pretrain" policy checkpoint provided by DexArt's official GitHub repository using a combined training set consisting of DexArt's original samples (sourced from PartNet-Mobility \cite{xiang2020sapien}) and additional samples (sourced from Objaverse \cite{deitke2024objaverse}) generated by \model. As a baseline, we fine-tune the policy for the same number of steps but only use the original dataset. During evaluation, the new policies are tested on both the original test set and the additional test samples generated by \model. We conducted 100 evaluations on the test set, where each evaluation included all test set objects and included randomization in position and rotation. The final success rate was obtained by averaging across all evaluations. Visualizations of the datasets used are provided in Figures \ref{fig:bk_train}, \ref{fig:lp_train}, and \ref{fig:ltest}.

\end{document}